\documentclass[letterpaper, 10 pt, conference]{ieeeconf}  

\IEEEoverridecommandlockouts                              

\usepackage{amsmath}
\usepackage{amssymb}
\usepackage{graphicx}
\usepackage[ruled,vlined]{algorithm2e}
\usepackage{breqn}
\usepackage{color}
\usepackage{ulem}
\usepackage{comment}
\usepackage{subcaption}
\usepackage{hyperref}

\title{\LARGE \bf
Jerk Constrained Velocity Planning for an Autonomous Vehicle: Linear Programming Approach
}

\author{Yutaka Shimizu$^{1, 2}$, Takamasa Horibe$^{2}$, Fumiya Watanabe$^{2}$ and Shinpei Kato$^{1, 2}$
\thanks{$^{1}$ Yutaka Shimizu and Shinpei Kato are with the Graduate School of Information Science and Technology, The University of Tokyo, 7-3-1 Hongo, Bunkyo-ku, Tokyo, 113-0033, Japan.
        {\tt\small \{yutaka.shimizu, shinpei.kato\}@pf.is.s.u-tokyo.ac.jp}}%
\thanks{$^{2}$ Yutaka Shimizu, Takamasa Horibe, Fumiya Watanabe and Shinpei Kato are with Tier IV, Inc., Jacom Building, 1-12-10 Kitashinagawa, Shinagawa-ku, Tokyo, 140-0001, Japan.
        {\tt\small \{takamasa.horibe, fumiya.watanabe, shinpei.kato\}@tier4.jp}}%
}

\begin{document}

\maketitle
\thispagestyle{empty}
\pagestyle{empty}

\begin{abstract}
Velocity Planning for self-driving vehicles in a complex environment is one of the most challenging tasks. It must satisfy the following three requirements: safety with regards to collisions; respect of the maximum velocity limits defined by the traffic rules; comfort of the passengers.
In order to achieve these goals, the jerk and dynamic objects should be considered, however, it makes the problem as complex as a non-convex optimization problem.
In this paper, we propose a linear programming (LP) based velocity planning method with jerk limit and obstacle avoidance constraints for an autonomous driving system.
To confirm the efficiency of the proposed method, a comparison is made with several optimization-based approaches, and we show that our method can generate a velocity profile which satisfies the aforementioned requirements more efficiently than the compared methods. In addition, we tested our algorithm on a real vehicle at a test field to validate the effectiveness of the proposed method.
\end{abstract}

\section{INTRODUCTION}

In years, feasibility of full autonomy in self-driving cars has been getting more clear thanks to the parallel developments in computing technologies and algorithms. 
Although hundreds of companies have raised generous funds, these initiatives still face many technical challenges in designing robust and optimal navigation algorithms that work for complex environments.
An ordinary traffic environment, that an autonomous vehicle is required to navigate, is generally complicated and unpredictable. Furthermore, the resulting problem structures are highly nonlinear and non-convex. Finding a solution to these problems on a real-time scale leaves developers with a few choices and compromising solutions, such as being content with a locally optimal solution rather than the global one. In some cases, ad-hoc and heuristic algorithms enhance the solution for the proposed systems, increasing the complexity, consequently the software maintenance and safe deployment. 

We address a similar problem structure in velocity planning which stipulates three core objectives; collision avoidance, keeping the vehicle within the regulated speed limits to follow the traffic rules and providing good comfort to the passengers.

Various algorithms and solutions have been proposed in the literature to achieve these goals. Among these, sampling-based methods are presented in~\cite{Ziegler2009Spatio,Lim2018Hierarchical,Hubmann2016AGeneric}. In these papers, the authors formulate the velocity planning problem as a graph search problem and seek the optimal solution,  making use of the graph search optimization methods. 
The resulting solutions are not smooth in general in their solutions. Therefore, the results require an additional smoothing procedure after the sampling~\cite{Meng2019Decoupled, Wenda2012AReal, Wei2017Spatially}, which increases the computational complexity and may give rise to difficulties in the implementation.
In an alternative direction, continuous optimization is often employed for the velocity planning problem due to its flexibility in defining cost functions and constraint equations. 
One major approach in the continuous optimization methods is using the nonlinear programming, which has one or more nonlinear and non-convex constraint equations~\cite{Zie2014Bertha,Qian2016Motion}.

When continuous optimization is used, velocity trajectory optimization problems can be parametrized by either time or arc-length (path-length parameterization)~\cite{Changliu2017Speed, frasch2013auto}.
The author in~\cite{zhou2020dliaps} applied time-parameterized quadratic programming optimization for velocity planning. In this approach, jerk related constraints can be described as linear form.
However, the maximum velocity can only be set for the entire path, not pointwise, which makes it difficult to define a specific location on the path to slow down before taking a curve. 
This requirement can be satisfied when arc-length parametrization is used. The pointwise constraints can be enforced at any point, however, jerk constraints for the passenger comfort introduce nonlinearity into the problem. 
In~\cite{zhang2018toward}, a pseudo-jerk is proposed as a substitute for the jerk, which is in a linear form. However, the pseudo-jerk cannot successfully approximate the actual jerk in a wide range of velocity.

In this paper, we propose a new approximation method to alleviate the issues we mentioned earlier. Our algorithm can handle jerk and collision constraints in the arc-parametrized velocity trajectory optimization. 
 In Section~\ref{sec:preliminary} and ~\ref{sec:preliminary}, we incorporate the obstacle avoidance constraints into the optimization by transforming them into the maximum velocity constraints analytically, then linearly approximate jerk constraint equations. 
As a result, the problem can be represented as an LP that can be solved in real-time to obtain global optimum solutions. 

We compared the proposed solution with a couple of optimization-based algorithms from the literature through several simulations in section~\ref{sec:numerical_experiment}. Numerical experiments show that our new approach satisfies all defined constraints of a given scenario with a minimal computational time than the others and successfully avoid collisions with dynamic obstacles. 
Furthermore, on-road experiments in section~\ref{sec:on_board_experiment} convince us that the proposed method can generate optimal velocity even in a real driving environment.

The main contribution of this paper can be summarized as follows.
\begin{itemize}

\item \textbf{Real-time dynamic obstacle constraints creator:} \newline
The proposed method can convert non-convex obstacle avoidance constraints into linear maximum velocity constraints quickly.
\item \textbf{Linear Jerk-Constrained Optimization:} \newline
The proposed method can solve a jerk-constrained velocity optimization problem parametrized by arc-length into a form of linear programming.
\item  \textbf{Numerical Simulations and Real-world Testings:} \newline
Numerical experiments and real-world testing prove that the proposed algorithm works better than any other existing method even in a complicated situation.

\end{itemize}

\section{PRELIMINARY} \label{sec:preliminary}
\subsection{Problem Formulation}\label{sec:problem_formulation}
First, we give a formal definition of an arc-length parametrized velocity planning problem within a scenario. The only assumption we make here is that the path shape is provided by a planner or map in advance.


Let the arc length along the path $s$ as the function of the time $t$, i.e., $s=s(t)$.

In this paper, we define the decision variables $b$ and $a$ for the velocity planning as follows:
\begin{equation} \label{eq:ba-continuous-definition}
b(s) := \dot{s}^2, \quad a(s) := \ddot{s}.
\end{equation}
Here, taking the derivative of the $b(s)$ derives the relation of these variables as
\begin{equation}
\frac{db}{ds} =\frac{d (\dot s^2) }{ds}= 2 \dot s \frac{d\dot s}{ds} = 2 \frac{ds}{dt} \frac{d\dot s}{ds} = 2 \ddot s = 2a
\end{equation}
which must be satisfied in the optimization problem.

By using the decision variables $b$ and $a$, the jerk $j$, which is taken as an indicator of ride comfort, can be expressed as follows:
\begin{equation}
j(s) := \dddot{s} = \dot{a} = \frac{da}{ds}\sqrt{b}.
\end{equation}
Furthermore, we must consider the visiting time of any specific locations on the path to incur the object avoidance constraints. The visiting time function of a location $s$ can be expressed as; 
\begin{equation}
    t(s) := \int_{0}^{s} b(\theta)^{-\frac{1}{2}} d\theta.
\end{equation}

The objective of the velocity planning problem we handle here is to increase the velocity as much as possible while considering velocity, acceleration, jerk, and obstacles avoidance constraints. The formal definition results in the following: 
\begin{subequations} \label{eq:problem-formulation}
\begin{align}
& \underset{b(s), a(s)}{\text{min}}
& & \int_{0}^{s} -b(\theta)d\theta \\
& \text{subject to} & &  \frac{db(s)}{ds} = 2a(s),  \\
& & &  v_{\mathrm{min}}^2(s) \leq b(s) \leq  v_{\mathrm{max}}^2(s), \\
& & &  a_{\mathrm{min}}(s) \leq a(s) \leq  a_{\mathrm{max}}(s), \\
\label{eq:problem-formulation-jerk}
& & &  j_{\mathrm{min}}(s) \leq \frac{da(s)}{ds}\sqrt{b(s)} \leq  j_{\mathrm{max}}(s), \\
\label{eq:problem-formulation-obs}
& & & T_{\mathrm{min}}(s) \leq \int_{0}^{s} b(\theta)^{-\frac{1}{2}} d\theta \leq T_{\mathrm{max}}(s),
\end{align}
\end{subequations}
where $v$, $a$, $j$ and $T$ represent velocity, acceleration, jerk, and passing time at $s$ respectively, and the ${\ast}_{\mathrm{max}}$, ${\ast}_{\mathrm{min}}$ subscripts represent the maximum and minimum values given as constraints. Note that \eqref{eq:problem-formulation-jerk} limits the value of the jerk, and \eqref{eq:problem-formulation-obs} is a constraint for obstacle avoidance. In this paper, we assume that $j_{\mathrm{min}} \leq 0$ and $0 \leq j_{\mathrm{max}}$.

For numerical computation, we discretize this problem. Let $N$ be a number of discretized trajectory points, and arc length at $i$th point are written as $s_i$. Then the discretized formulation of \eqref{eq:problem-formulation} becomes,
\begin{subequations} \label{eq:discretized-problem-formulation}
\begin{align}
& \underset{\mathbf{b}, \mathbf{a} \in \mathbb{R}^N}{\text{min}}
\label{eq:discretized-problem-formulation-cost}
& & \sum_{k=1}^{N} -b_k \\
& \text{subject to} & &  b_{i+1} - b_{i} = 2a_i\left( s_{i+1} - s_{i} \right),  \\
\label{eq:discretized-velocity-constraint}
& & &  v_{\mathrm{min}, i}^2 \leq b_i \leq  v_{\mathrm{max}, i}^2, \\
& & &  a_{\mathrm{min}, i} \leq a_i \leq  a_{\mathrm{max}, i}, \\
\label{eq:discretized-problem-formulation-jerk}
& & &  j_{\mathrm{min}, i} \leq \left(\frac{a_{i+1} - a_i}{s_{i+1}-s_i}\right) \sqrt{b_i} \leq  j_{\mathrm{max}, i}, \\
\label{eq:discretized-problem-formulation-obs}
& & & T_{\mathrm{min},i} \leq \sum_{k=1}^{i-1} \frac{(s_{k+1}-s_k)}{\sqrt{b_k}} \leq T_{\mathrm{max},i}, \\
& & & (i = 1, \ldots, N), \nonumber
\end{align}
\end{subequations}
where $b_i$ and $a_i$ are $b(s_i)$ and $a(s_i)$. Moreover, vector $\mathbf{b}$ and $\mathbf{a}$ are defined as $\mathbf{b}:=[b_1, b_2, \cdots, b_N]^{\sf T}$ and $\mathbf{a}:=[a_1, a_2, \cdots, a_N]^{\sf T}$.
However, since the constraints,  given by \eqref{eq:discretized-problem-formulation-jerk} are non-convex and ~\eqref{eq:discretized-problem-formulation-obs} are nonlinear, it is challenging to find the optimal solution of the problem~\eqref{eq:discretized-problem-formulation} in real-time, especially when the number of discretization points $N$ is large.

\subsection{Pseudo-Jerk}\label{sec:review}
Since the jerk constraint~\eqref{eq:discretized-problem-formulation-jerk} cannot be converted to convex form, \cite{zhang2018toward} proposed a new indicator called pseudo-jerk, which is defined as:
\begin{equation}
    j^{\mathrm{pseudo}}_i := \frac{j_i}{\sqrt{b_i}} = \frac{a_{i+1}-a_i}{s_{i+1}-s_i}  \label{eq:pjerk-definition}
\end{equation}
where $j_i$ is the jerk at the $i$th point.
In other words, pseudo-jerk is the first derivative of the acceleration with respect to the arc length $s$.
By using the pseudo-jerk, they remove the jerk constraint~\eqref{eq:discretized-problem-formulation-jerk} and add the smoothing cost.
As a result, the overall problem becomes
\begin{subequations} \label{eq:pjerk-formulation}
\begin{align}
& \underset{\mathbf{b}, \mathbf{a} \in \mathbb{R}^N}{\text{min}}
\label{eq:pjerk-formulation-cost}
& & \sum_{k=1}^{N} -b_k + w_{\mathrm{smooth}} \sum_{k=1}^{N-1} \left( j_k^{\mathrm{pseudo}} \right)^2 \\
& \text{subject to} & &  b_{i+1} - b_{i} = 2a_i\left( s_{i+1} - s_{i} \right),  \\
\label{eq:pjerk-formulation-velocity}
& & &  v_{\mathrm{min}, i}^2 \leq b_i \leq  v_{\mathrm{max}, i}^2, \\
& & &  a_{\mathrm{min}, i} \leq a_i \leq  a_{\mathrm{max}, i},\\
& & & T_{\mathrm{min},i} \leq \sum_{k=1}^{i-1} \frac{(s_{k+1}-s_k)}{\sqrt{b_k}} \leq T_{\mathrm{max},i}, \\
& & & (i = 1, \ldots, N),
\end{align}
\end{subequations}
where $w_{\mathrm{smooth}}$ is a hyper-parameter to tune smoothness. Problem~\eqref{eq:pjerk-formulation} can be solved by using the quadratic programming (QP) technique.

Even thought this approach can generate a smooth velocity profile, the pseudo-jerk~\eqref{eq:pjerk-definition} cannot successfully approximate the actual jerk in a wide velocity range (e.g. at high velocity, the pseudo jerk is lower than the actual jerk, and conversely, it is infinite at zero velocity), thus, there is still room to improve the driving comfort.

\section{METHODOLOGY} \label{sec:methodology}
Our approach is to relax the non-convex velocity planning problem to convexify the jerk inequalities by using the approximated velocity profile. This can be done in two steps. In the first step, we design and apply a velocity limit filter that converts the obstacle avoidance constraints into a maximum velocity constraint. After that, we apply a forward-backward jerk filter to improve the approximation accuracy and linearize jerk constraints.

\subsection{Obstacle Avoidance Constraint} \label{section:obstacle-filter}
Instead of using the obstacle avoidance constraint formulated in~\eqref{eq:discretized-problem-formulation-obs}, we introduce a new approach to avoid an obstacle by modifying a maximum velocity constraint.

In this paper, it is assumed that the trajectory of a dynamic obstacle is represented as a continuous linear function of the velocity of the obstacle $v_{\mathrm{obs}}$, such that
\begin{equation}
s_{\mathrm{obs}}(t) = v_{\mathrm{obs}}(t - t_{\mathrm{in}}) + s_{\mathrm{in}}, \quad t \in [t_{\mathrm{in}}, t_{\mathrm{out}}],
\end{equation}
where $s_{\mathrm{in}}$ denotes the arc length on the ego path which the obstacle cuts in, and $t_{\mathrm{in}}$ and $t_{\mathrm{out}}$ denote the cut in time and the cut out time of the obstacle, respectively.
Let us define two positive values $d_{\alpha}$ and $d_{\beta}$, such that $d_{\alpha} \geq d_{\beta}$. First, the ego vehicle maintains the same maximum velocity as the original maximum velocity until the distance to the dynamic obstacle becomes $d_{\alpha}$. When the distance to the obstacle becomes less than $d_{\alpha}$, the velocity of the vehicle is adjusted as follows.
\begin{equation}
    v_{\alpha} := \frac{s_{\mathrm{out}} - s_{\alpha}}{t_{\mathrm{out}} - t_{\alpha}},
\end{equation}
where $s_{\alpha}$ denotes the arc length at which the distance between the ego vehicle and the obstacle is $d_{\alpha}$. In addition, $t_{\alpha}$ is the time at which the ego vehicle reaches $s_{\alpha}$ with the maximum velocity, i.e.,
\begin{equation}
t_{\alpha}(s_{\alpha}) := \int_0^{s_{\alpha}} \frac{1}{v_{\max}(\theta)} d\theta,
\end{equation}
and $s_{\mathrm{out}}$ denotes the arc length at which the obstacle cuts out the ego vehicle path, i.e.,
\begin{equation}
    s_{\mathrm{out}} := s_{\mathrm{obs}}(t_{\mathrm{out}}).
\end{equation}
Furthermore, when the distance to the dynamic obstacle reaches $d_{\beta}$, the maximum velocity is switched to $v_{\beta}$:
\begin{equation}
    v_{\beta} := v_{\mathrm{obs}}.
\end{equation}
The entire algorithm is described in Algorithm~\ref{alg:obstacle-filter}.
Note that $\mathbf{v}_{\mathrm{max}}$ denotes the original maximum velocity, which is $\mathbf{v}_{\mathrm{max}} := [v_{\mathrm{max}, 1}, \ldots, v_{\mathrm{max}, N}]^{\sf T}$ and, $\mathbf{s_{\mathrm{ego}}}$ denotes the arc length along the ego path i.e. $\mathbf{s_{\mathrm{ego}}} := [s_1, \ldots, s_N]^{\sf T}$. In addition, $t_{\mathrm{range}}$ indicates the threshold for the time difference between the obstacles and the ego vehicle.

\begin{algorithm}[t]
    \SetAlgoLined
    \SetNoFillComment
    \SetKwData{Left}{left}\SetKwData{This}{this}\SetKwData{Up}{up}
    \SetKwFunction{Union}{Union}\SetKwFunction{FindCompress}{FindCompress}
    \SetKwInOut{Input}{input}\SetKwInOut{Output}{output}
    \Input{Obstacle Trajectory $\left(\mathbf{s_{\mathrm{obs}}},\mathbf{t_{\mathrm{obs}}}\right)$\\
    Maximum Velocity Profile $(\mathbf{s_{\mathrm{ego}}}, \mathbf{v}_{\mathrm{max}})$}
    \Output{Updated Maximum Velocity $\mathbf{\hat{v}}_{\mathrm{max}}$}
    \For{$k=1$ \KwTo $N - 1$}
    {
        Find $(s_{\mathrm{obs}}, t_{\mathrm{obs}})$ which is the nearest of $\left(s_k,(s_{k+1} - s_k )/v_k \right)$;\\
        $ds$ = $|s_k-s_{\mathrm{obs}}|$; \\
        $dt$ = $|t_k-t_{\mathrm{obs}}|$; \\
        \uIf{$ds \leq d_{\beta} \And dt \leq t_{\mathrm{range}}$}
        {
            $\hat{v}_{\mathrm{max}, k}$ $\leftarrow$ $v_{\beta}$; \\
        }
        \uElseIf{$ds \leq d_{\alpha} \And dt \leq t_{\mathrm{range}}$}
        {
            $\hat{v}_{\mathrm{max}, k}$ $\leftarrow$ $v_{\alpha}$; \\
        }
        \Else
        {
            $\hat{v}_{\mathrm{max}, k}$ $\leftarrow$ $v_{\mathrm{max}, k}$; \\
        }
    }
    \Return $\mathbf{\hat{v}_{max}}$
    \caption{Velocity Limit Filter}
    \label{alg:obstacle-filter}
\end{algorithm}

\begin{figure}[t]
    \centering
    \includegraphics[scale=0.2]{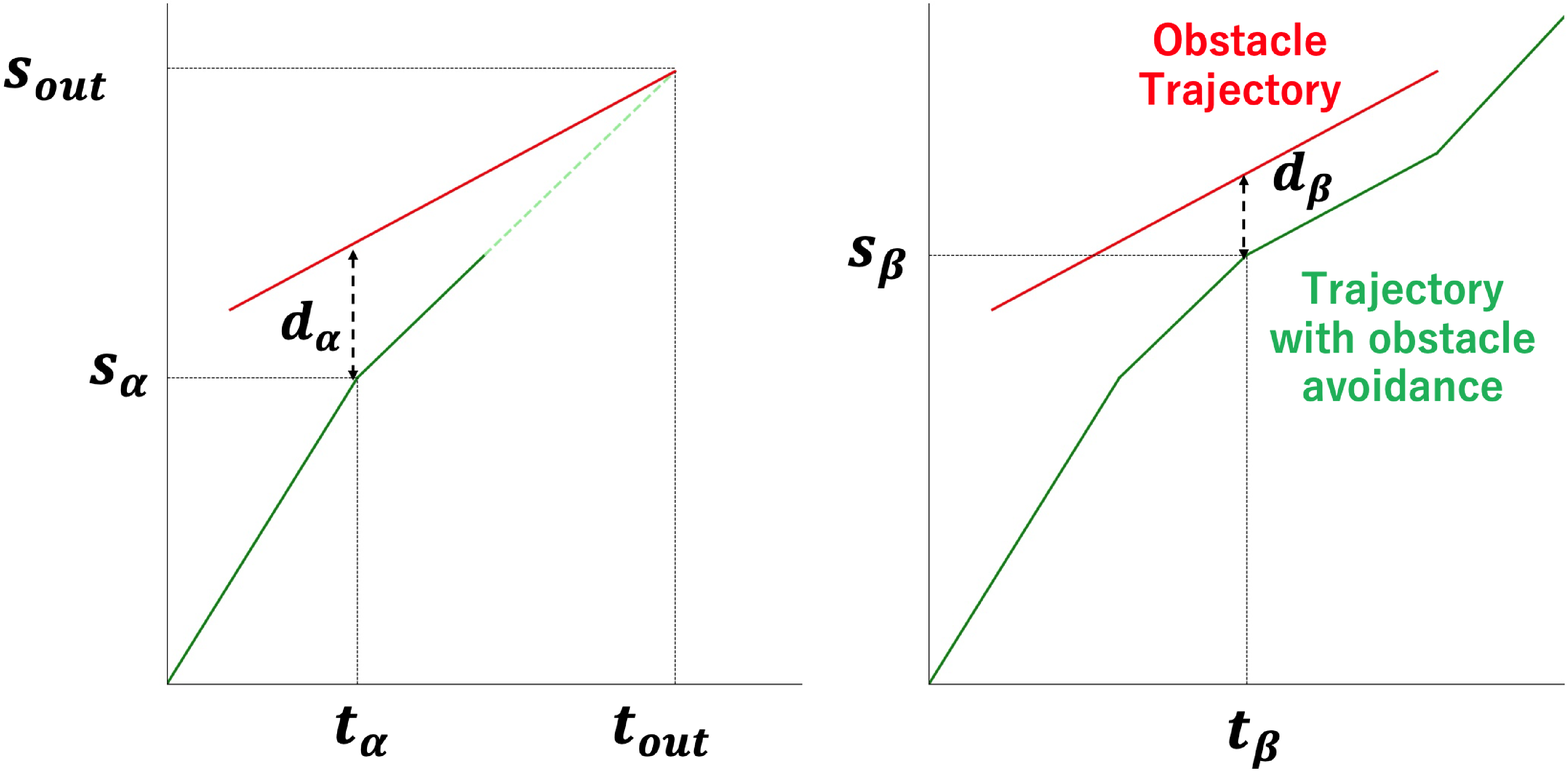}
    \caption{Example of the velocity limit filter for obstacle avoidance. The red line and the green line denote the trajectory of the dynamic obstacle and the trajectory with the constrained maximum velocity, respectively.}
    \label{fig:constrained-maximum-velocity}
\end{figure}

We depict an example of these procedures in  Fig.\ref{fig:constrained-maximum-velocity}.
The ego vehicle starts travelling at the given maximum velocity, and the maximum velocity is switched to $v_{\alpha}$ at $s_{\alpha}$. Then, when the distance to the obstacle becomes $d_{\beta}$, the maximum velocity is set to $v_{\beta}$.
After the dynamic obstacle disappears from the ego vehicle trajectory, the maximum velocity is set back to its original value.

\subsection{Jerk Filter} \label{section:jerk-filter}
In this section, we describe how to deal with the nonlinear jerk constraint~\eqref{eq:problem-formulation-jerk}.
What makes the problem nonlinear in equation~\eqref{eq:problem-formulation-jerk} is the bi-linear term $a_i\sqrt{b_i}$. This means that if we can fix $\sqrt{b}$, the equation becomes linear. Our base idea is to make a rough estimate of an optimized velocity profile and replace $\sqrt{b}$ with it during optimization so that ~\eqref{eq:problem-formulation-jerk} becomes linear. That estimate is made by the forward-backward jerk filter described below.

\begin{figure}[t]
    \centering
    \includegraphics[scale=0.2]{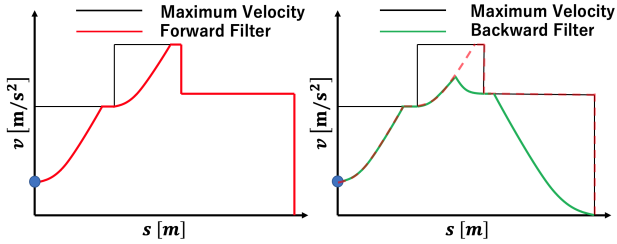}
    \caption{The result of jerk filtering. The left and right figures show the forward and backward filters, respectively.}
    \label{fig:jerk-filter}
\end{figure}

To estimate the optimized velocity in advance, we applied two filters for a given maximum velocity profile: forward jerk filtering and backward jerk filtering.
In the forward step, we update the velocity with the maximum acceleration and jerk in the following way.
\begin{subequations} \label{eq:jerk-filter-forward}
\begin{align}
    \label{eq:jerk-filter-forward-time}
    dt &= \frac{ds}{v_{f,k}} \\
    a_{k+1} &= \min\left(a_{\mathrm{max}}, a_k + j_{\mathrm{max}} dt\right) \\
    v_{f,k+1} &= \max\left(v_{\mathrm{max}}, v_{f,k} + a_{k+1}  dt\right)
\end{align}
\end{subequations}
where $v_{f,k}$ denotes the filtered velocity at $s_k$. 
This process generates the velocity profile as shown on the left in Fig.~\ref{fig:jerk-filter}.
In the backward step, the same filtering logic is applied to the opposite direction (from $k=N-1$ to $2$) and it generates a velocity profile shown on the right in Fig.~\ref{fig:jerk-filter} as a green line. 
Although there are discontinuous accelerations in the filtered velocity, it can be used as a rough estimate of an optimization result. 
The details of this algorithm are described in Algorithm \ref{alg:jerk-filter}.

Let $\mathbf{v}_{{f}}$ be the velocity generated by the jerk filter. Replacing $\sqrt{b_i}$ in~\eqref{eq:problem-formulation-jerk} with $v_{f,i}$, we can approximate the jerk constraint inequality as follows:
\begin{equation}\label{eq:filter_velocity_jerk_constraint}
j_{\mathrm{min}, i} \leq \left(\frac{a_{i+1} - a_i}{s_{i+1}-s_i}\right) v_{{f}, i} \leq  j_{\mathrm{max}, i}.
\end{equation}
Note that this equation is linear with the decision variables $a_i$ and $a_{i+1}$.

\begin{algorithm}[t]
    \SetAlgoLined
    \SetNoFillComment
    \SetKwData{Left}{left}\SetKwData{This}{this}\SetKwData{Up}{up}
    \SetKwFunction{Union}{Union}\SetKwFunction{FindCompress}{FindCompress}
    \SetKwInOut{Input}{input}\SetKwInOut{Output}{output}
    \Input{Maximum Velocity Profile $\left(\mathbf{s_{\mathrm{ego}}},\mathbf{v_{\mathrm{max}}}\right)$\\
    Acceleration Limit Profile $\left(\mathbf{s_{\mathrm{ego}}},\mathbf{a_{\mathrm{min}}}, \mathbf{a_{\mathrm{max}}}\right)$\\
    Jerk Limit Profile $\left(\mathbf{s_{\mathrm{ego}}},\mathbf{j_{\mathrm{min}}},\mathbf{j_{\mathrm{max}}}\right)$}
    \Output{Filtered Velocity $\mathbf{v_{f}}$}
    $v_{\mathrm{forward},1} = v_{\mathrm{max}, 1}$ \\
    $a_{\mathrm{forward},1} = a_{\mathrm{max}, 1}$ \\
    \For{$k=2$ \KwTo $N$}
    {
        $dt = (s_k - s_{k-1})/v_{\mathrm{forward},k-1}$ \\
        $a_{\mathrm{forward},k} = \min\left(a_{\mathrm{max}, k},\ a_{\mathrm{forward},k-1} + j_{\mathrm{max}, k-1} dt\right)$ \\
        $v_{\mathrm{forward},k} = \max\left(v_{\mathrm{max}, k},\ v_{\mathrm{forward},k-1} + a_{\mathrm{forward},k-1}  dt\right)$ \\
    }
    $v_{\mathrm{backward},N} = v_{\mathrm{forward}, N}$ \\
    $a_{\mathrm{backward},N} = 0$ \\
    \For{$k=N-1$ \KwTo $2$}
    {
        $dt = (s_k - s_{k+1})/v_{\mathrm{backward},k+1}$ \\
        $a_{\mathrm{backward}, k} = \min\left(a_{\mathrm{min},k},\ a_{\mathrm{backward}, k+1} - j_{\mathrm{min}, k} dt\right)$ \\
        $v_{\mathrm{backward},k} = \max\left(v_{\mathrm{max}, k},\ v_{\mathrm{backward}, k+1} - a_{\mathrm{backward}, k+1}  dt\right)$ \\
    }
    \Return $\mathbf{v_{f}} = \mathbf{v_{\mathrm{backward}}}$
    \caption{Jerk Filter}
    \label{alg:jerk-filter}
\end{algorithm}

Let us consider the accuracy of this approximation. Since $\mathbf{v}_{f}$ is the result of using the maximum jerk and acceleration, the optimization result will not exceed the filtered velocity. In other words, we can say that $\sqrt{b_i} \leq {v}_{f,i}$. Since the cost function of the optimization is designed to make $\mathbf{b}$ as large as possible, the optimized velocity profile $\mathbf{b}$ will not deviate significantly from the filtered velocity $\mathbf{v}_{f}$. Therefore, this jerk constraint can be maintained with a certain level of accuracy. This will be shown in the numerical experiment section later.

Furthermore, rewriting \eqref{eq:filter_velocity_jerk_constraint} using an actual jerk $(a_{i+1} - a_{i})/(s_{i+1} - s_i)\sqrt{b_i}$, we get
\begin{equation}
\label{eq:jerk-lim-estimate}
    j_{\mathrm{min}, i} \frac{\sqrt{b_i}}{v_{f, i}} \leq \left(\frac{a_{i+1} - a_i}{s_{i+1}-s_i}\right) \sqrt{b_i} \leq  j_{\mathrm{max}, i} \frac{\sqrt{b_i}}{v_{f, i}}
\end{equation}
Since $0\leq \sqrt{b_i} \leq {v}_{f,i}$, the optimized jerk will not exceed the given jerk limit, which is $j_{\mathrm{min}, i} \leq j_i \leq j_{\mathrm{max}, i}$. Note that if $v_{f,i} = 0$, then $\sqrt{b_i}=0$, thus the jerk is 0.

\subsection{Linear Programming Formulation} \label{section:linear-programming}
In this section, we reformulate \eqref{eq:discretized-problem-formulation} by utilizing the results of Section~\ref{section:obstacle-filter} and \ref{section:jerk-filter}.

As stated in Section~\ref{section:obstacle-filter}, since the obstacle avoidance constraint is converted to the maximum velocity constraint, the collision avoidance constraints \eqref{eq:discretized-problem-formulation-obs} can be ignored in the optimization problem.
Therefore, the optimization problem can be written as follows:
\begin{subequations} \label{eq:lp-optimization}
\begin{align}
& \underset{\mathbf{b}, \mathbf{a} \in \mathbb{R}^N}{\text{min}}
& & \sum_{k=1}^{N}-b_k \\
& \text{subject to} & &  b_{i+1} - b_{i} = 2a_i\left( s_{i+1} - s_{i} \right),  \\
\label{eq:lp-optimization-velocity}
& & &  v_{\mathrm{min}, i}^2 \leq b_i \leq  v_{f, i}^2, \\
& & &  a_{\mathrm{min}, i} \leq a_i \leq  a_{\mathrm{max}, i}, \\
& & &  j_{\mathrm{min}, i}  \leq \left(\frac{a_{i+1} - a_i}{s_{i+1}-s_i}\right)v_{f,i} \leq  j_{\mathrm{max}, i}.
\end{align}
\end{subequations}
This is a linear programming that can be solved efficiently. 

\section{NUMERICAL EXPERIMENT} \label{sec:numerical_experiment}
\subsection{Situation Settings}
In this section, we evaluate the proposed LP-based approach and compare it with the Non-Convex approach for \eqref{eq:discretized-problem-formulation} and the QP-based pseudo-jerk approach~\cite{zhang2018toward} for \eqref{eq:pjerk-formulation}. We use Gurobi optimization solver\cite{gurobi} to solve LP and QP problems, and the NLOPT\cite{NLOPT} to solve a Non-Convex optimization problem. The algorithms are implemented in C++ and performed on a Macbook-Pro 2019 with 2.40GHz Intel Core i9 CPU. 
Our test code is available here\footnote{\label{footnote1}\url{https://github.com/pflab-ut/jerk_optimal_velocity_planning}}.

We set the sampling size of the path as $N=300$ and the distance of the sampled path point as $ds = 0.1$. 
The acceleration limits and the jerk limits are set to $j_{\mathrm{max}, i} = 0.8$, $j_{\mathrm{min}, i} = -0.8$, $a_{\mathrm{max}, i} = 1.0$, $a_{\mathrm{max}, i} = -1.0$, for $i = 2, \ldots, N$.
For initial values of velocity and acceleration, $v_{\mathrm{max}, 1} = 0.5$, $v_{\mathrm{min}, 1} = 0.5$, $a_{\mathrm{max}, 1} = 0.0$, $a_{\mathrm{min}, 1} = 0.0$.

\subsection{Filtered Velocity Profile}
\begin{figure}[t]
    \centering
    \includegraphics[scale=0.45]{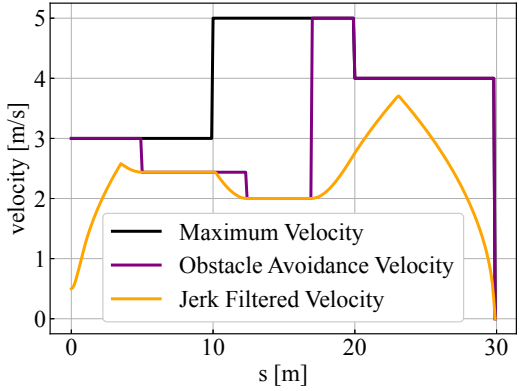}
    \caption{The result of velocity limit filtering and jerk filtering.}
    \label{fig:experiment-filter}
\end{figure}

First, we show the result of velocity profile filtering proposed in Section ~\ref{section:obstacle-filter} and \ref{section:jerk-filter}.
The filtered velocity profile is drawn in the Fig.~\ref{fig:experiment-filter}.
In this figure, the black line denotes the original maximum velocity profile, the purple line denotes the maximum velocity profile for obstacle avoidance and the orange line denotes the jerk filtered maximum velocity profile. These speed profiles are used in the proposed LP formulation.

\subsection{Planning Result}
\begin{figure}[t]
    \centering
    \includegraphics[scale=0.13]{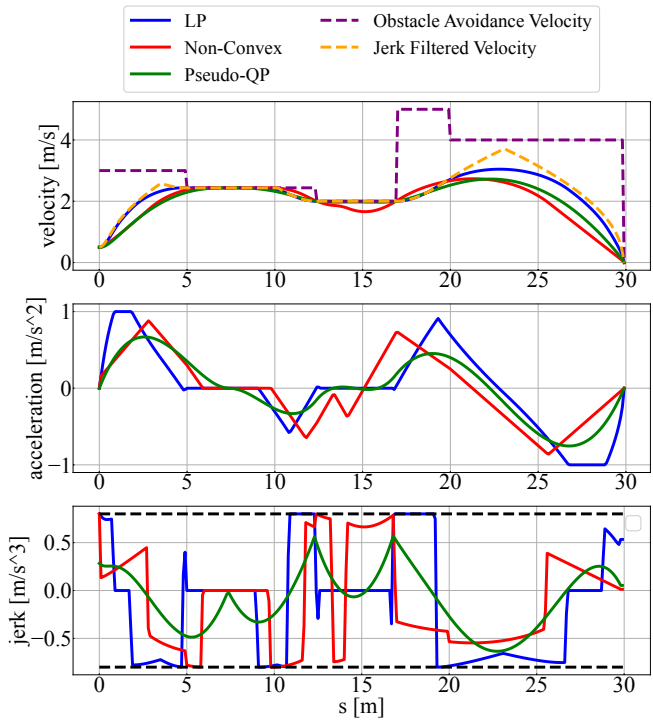}
    \caption{The comparison between our proposed method (blue) for~\eqref{eq:lp-optimization} and QP (green) for~\eqref{eq:pjerk-formulation} and nonlinear optimization (red) for~\eqref{eq:problem-formulation}. The dashed lines on the top plot denote the velocity limits generated by the proposed pre-filters. The dashed black line on the bottom plot is the jerk limit.}
    \label{fig:experiment-optimization-results}
\end{figure}

We show the result of the LP, QP, and Non-Convex velocity planning in Fig.~\ref{fig:experiment-optimization-results} for a scenario (additional examples can be found here\textsuperscript{\ref{footnote1}}).
It shows the resulting velocity profiles, acceleration and jerk from top to bottom. The purple and orange dashed line in this picture denotes the maximum velocity profile for obstacle avoidance and the jerk filtered maximum velocity profile, respectively. 
It is seen that the LP and Non-Convex approach satisfies all the constraints defined for velocity, acceleration and jerk properly. That is to say, although the proposed method solves an approximation of the original problem, the generated velocity profile does not violate the original constraints.
In the pseudo-jerk approach, since there is no jerk constraint defined in the constraint equations, we tuned the weight term of the jerk in the objective function in such a way that jerk does not exceed the jerk limit  ($w_{\mathrm{smooth}}=500$ in this case). 
However, since the pseudo-jerk does not properly represent the jerk, the jerk is too small at some points resulting in poor acceleration and deceleration profile. For instance, the initial acceleration of the pseudo-jerk approach is smaller than proposed method because of the small jerk values. 
When looking at our proposed method, the optimal jerk profile tends to stay close to the maximum and minimum limits compared to the pseudo-jerk approach. This means that acceleration and deceleration are performed more efficiently within the constraints.


The computational time of the proposed method is 5.10 ms (1.5 \% for the pre-filterings, and 98.5 \% for the optimization), while the pseudo-jerk approach is 7.21 ms, and the non-convex approach is 3463 ms.

\section{On-road Experiment} \label{sec:on_board_experiment}
\subsection{Setup}
In order to demonstrate the effectiveness of the proposed method in a complex real-world environment, we conducted several experiments using our test vehicle, at a test field in the University of Tokyo test facilities. We give details of the sensor and computer configuration of the test vehicle in Table \ref{tab:test_vehicle_configuration}. Our test vehicle is running on Autoware\cite{kato2015open, kato2018autoware}, and since it uses OSQP\cite{osqp} as a default solver, we also use OSQP to solve optimization problems for the jerk-constrained velocity planning problem for on-road experiments.

\begin{table}
  \begin{center}
    \caption{Test vehicle information}
    \label{tab:test_vehicle_configuration}
    \begin{tabular}{|c|c|c|} 
      \hline
      \textbf{Component} & \textbf{Company} & \textbf{Type} \\
      \hline
      Lidar(Roof Top) & Velodyne & VLS128 Alpha Prime \\
      \hline
      Lidar(Roof Side) & Velodyne & VLS16 \\
      \hline
      Lidar(Roof Rear) & Velodyne & VLS16 \\
      \hline
      Lidar(Bumper Front) & Livox & Livox Horizon\\
      \hline
      Radar & Continental & - \\
      \hline
      Camera(Object Recognition) & FLIR & Blackfly S \\
      \hline
      Camera(Signal Recognition) & Leopard & IMX490 \\
      \hline
      PC CPU & Intel &  Core i7-8700 3.20GHz \\
      \hline
      PC GPU & Nvidia & Geforce GTX 1660\\
      \hline
      PC Memory & - & 32 GB\\
      \hline
      OS & - & Ubuntu 18.04 \\
      \hline
    \end{tabular}
  \end{center}
\end{table}

\subsection{Experiment Scenarios}
In this study, we conducted real-world tests for three different scenarios; stopping for a static obstacle,  vehicle following and collision avoidance with a cut-in obstacle. In each scenario, We use the following distance functions for obstacle avoidance constraint:
\begin{subequations} \label{eq:obstacle-filter-value}
\begin{gather}
    d_\alpha = d_\beta, \\
    d_\beta =\min(v_{\mathrm{obs}} t_{\mathrm{safe}}, \ d_{\mathrm{min}}) + d_{\mathrm{margin}},
\end{gather}
\end{subequations}
where $v_{\mathrm{obs}}$ denotes the velocity of the obstacle. Since we assume that an obstacle travels in a constant velocity over the planning horizon, $v_{\mathrm{obs}}$ would be the current velocity of the obstacle obtained from the tracking module in Autoware. We set $t_{\mathrm{safe}} = 2.0$, $d_{\mathrm{min}}=3.0$ and $ d_{\mathrm{margin}}=2.0$ in this experiment. For optimization, jerk and acceleration limits are set to $j_{\mathrm{max}, i} = 1.0 \ \mathrm{m/s^3}$, $j_{\mathrm{min}, i} = -0.5 \ \mathrm{m/s^3}$, $a_{\mathrm{max}, i} = 1.0 \ \mathrm{m/s^2}$, $a_{\mathrm{max}, i} = -0.5 \ \mathrm{m/s^3}$.

\subsubsection{Stopping}
In this test scenario, we put a static obstacle on the trajectory ahead of the ego vehicle and check if the vehicle can stop in front of the obstacle with the specified distance. 

\subsubsection{Vehicle Following}
Next, we conduct an experiment to follow a dynamic obstacle in front of the ego vehicle. In this scenario, the maximum speed for the ego vehicle is set to $15\ \mathrm{km/h}$, and the obstacle runs at a constant speed of $10\ \mathrm{km/h}$. In this case, we also evaluate that the proposed algorithm can decrease before the curve, and to achieve this goal we impose the following maximum velocity constraint for the ego vehicle. 

\begin{equation} \label{eq:experiment_curve_decrease}
    v_{\mathrm{max}} = \min \left(v_{\mathrm{max, road}}, \sqrt{\frac{a_{\mathrm{max, lateral}}}{\kappa}}\right),
\end{equation}
where $v_{\mathrm{max, road}}$ is the maximum velocity of the road, which is $30\ \mathrm{km/h}$ in this case, $a_{\mathrm{max,lateral}}$ is the maximum lateral acceleration value and $\kappa$ is the curvature value at each position on the trajectory. We set the lateral acceleration limit $a_{\mathrm{max, lateral}} = 0.8\ \mathrm{m/s^2}$ to limit the tangential velocity of the ego vehicle while taking a curve on the trajectory.
Note that, we can get maximum velocity by using ~\eqref{eq:experiment_curve_decrease} before the obstacle constraint and jerk filter calculations since the curvature of the path can be obtained as soon as the path is generated by the path planner. 

\subsubsection{Cut-in Vehicle}
In the final experiment scenario, we test the response of the proposed method when an obstacle cuts-in the ego trajectory. 

\subsection{Result}
We put the results of the on-board experiment in our Github page due to the lack of the space. From the video and the plot, It is clear that the optimization results respects the constraints as planned and avoid the collision with dynamic obstacles. the vehicle slows down while taking a curve due to the lateral acceleration constraint put in the problem. This is a piece of clear evidence showing that the proposed method can also be used for the problems where curved paths are involved.
More details about this experiment can be found from our videos, and entire code will be released on \url{https://github.com/tier4/AutowareArchitectureProposal.proj}.


\section{CONCLUSION} \label{sec:conclusion}
In this paper, we have proposed an algorithm for the jerk constrained velocity optimization problem. 
First, we formulate the obstacle avoidance constraint as the maximum velocity constraint and then, we converted the non-convex jerk constraint in the optimization problem to a linear constraint by taking advantage of the proposed jerk filters, which generate approximated optimal velocity.
Finally, the velocity planning problem is formulated as linear programming, and the optimal solution  can be found efficiently
Furthermore, we showed that the proposed LP solution with jerk filters achieves the fastest computations with better accuracy than the compared peers in the real vehicle experiments and computational simulations. 




\section*{ACKNOWLEDGMENT}
This work was supported by JST CREST Grant Number JPMJCR19F3, Japan.


\bibliographystyle{IEEEtran}
\bibliography{reference}

\end{document}